\documentclass[journal]{IEEEtran}
\usepackage{times}
\usepackage{epsfig}
\usepackage{graphicx}
\usepackage{amsmath}
\usepackage{amssymb}
\usepackage{epstopdf}
\usepackage{subfigure}
\usepackage{comment}
\usepackage[linesnumbered,ruled,lined]{algorithm2e}
\usepackage{algpseudocode}
\usepackage{color}
\usepackage{booktabs}
\usepackage{multirow}

\ifCLASSINFOpdf
\else
\fi
\begin{document}
%
\title{Affine-modeled video extraction \\ from a single motion blurred image}
%
%
%

\author{Daoyu~Li,
        Liheng~Bian*,
        and~Jun~Zhang,
\thanks{D. Li, L. Bian and J. Zhang are with the School of Information and Electronics \& Advanced Research Institute of Multidisciplinary Science, Beijing Institute of Technology, Beijing 100081, China. Corresponding e-mail: bian@bit.edu.cn.}}
\maketitle

\begin{abstract}
A motion-blurred image is the temporal average of multiple sharp frames over the exposure time. Recovering these sharp video frames from a single blurred image is nontrivial, due to not only its strong ill-posedness, but also various types of complex motion in reality such as rotation and motion in depth. In this work, we report a generalized video extraction method using the affine motion modeling, enabling to tackle multiple types of complex motion and their mixing. In its workflow, the moving objects are first segemented in the alpha channel. This allows separate recovery of different objects with different motion. Then,  we reduce the variable space by modeling each video clip as a series of affine transformations of a reference frame, and introduce the $l0$-norm total variation regularization to attenuate the ringing artifact. The differentiable affine operators are employed to realize gradient-descent optimization of the affine model, which follows a novel coarse-to-fine strategy to further reduce artifacts. As a result, both the affine parameters and sharp reference image are retrieved. They are finally input into stepwise affine transformation to recover the sharp video frames. The stepwise retrieval maintains the nature to bypass the frame order ambiguity. Experiments on both public datasets and real captured data validate the state-of-the-art performance of the reported technique.
\end{abstract}

\begin{IEEEkeywords}
video extraction, affine model, total variation, coarse-to-fine
\end{IEEEkeywords}

%
\IEEEpeerreviewmaketitle

\section{Introduction}
%
%
%
%
\IEEEPARstart{W}{hen} capturing images and videos, the relative motion between camera and object leads to motion blur that degrades image quality \cite{tiwari2013review}. A motion blurred image is the temporal average of underlying multiple sharp frames over the exposure time \cite{balakrishnan2019visual,Jin_2018_CVPR,purohit2019bringing}. The blurred image contains both the texture and motion information of the moving object. Recovering the sharp latent texture from the blurred image is an ill-posed task, and has been extensively investigated as deblurring \cite{tiwari2013review,ruiz2015variational}. Generally, these deblurring methods treat the motion blurred image as a convolution of a sharp image and blur kernel, and restore the latent image and blur kernel by various algorithms. Recently, the learning-based deblurring has been studied to offer higher reconstrcution accuracy and fast inference speed \cite{nah2019ntire}.

Despite of the extensive deblurring studies, most of the methods fail when motion trajectory is out of the image plane \cite{whyte2012non,zheng2013forward}. Moreover, the deblurring task treats motion blur as a degradation cause and tries to get rid of it. However, motion blur offers the object's dynamic information over exposure time, which is indispensable in multiple applications that require high-speed imaging \cite{balakrishnan2019visual,dillavou2019virtual,Jin_2018_CVPR,purohit2019bringing}. Based on this underlying observation, we engage to tackle the video extraction task, which retrieves a sharp video clip instead of an image from a single blurred image.

\begin{figure}[t]
	\centering
	\includegraphics[width=1\linewidth]{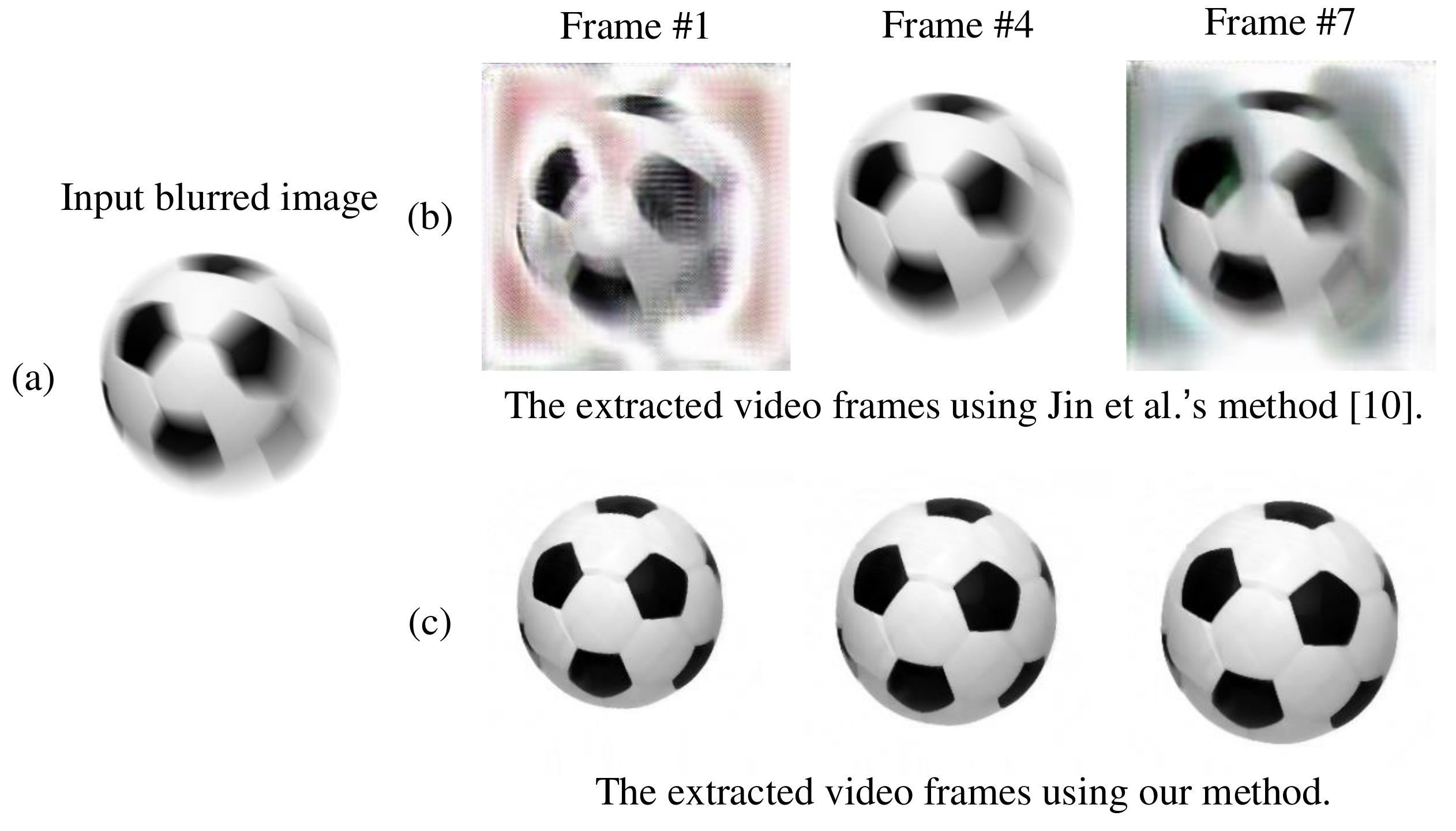}
	\caption{Video extraction from a single blurred image. Different from the conventional deblurring techniques that recover one sharp image from a blurred image, video extraction aims to retrieve a high-speed video clip. Our method enables to tackle 3D motion whose motion trajectory is out of the image plane. As a demonstration, the blurred image in (a) contains both motion in depth and rotation. Compared with the reconstruction results in (b) by Jin et al. \cite{Jin_2018_CVPR}, our method successfully recovers the sharp video clip as shown in (c). Note that the scale change of the football object dues to motion in depth.}
	\label{fig:demo}
\end{figure}

The video extraction task is nontrivial and challenging due to the following three reasons: (1) The problem is much more ill-posed than deblurring because there are more images to be recovered. (2) There exists the frame order ambiguity (chronological uncertainty) \cite{Jin_2018_CVPR}, which arises from the fact that the average of multiple images in any order would result in the same blurred image. (3) There exist multiple types of motion in reality such as rotation and motion in depth, which are more complex than the common translation motion.

Most recently, there appears a few studies trying to tackle the video extraction problem \cite{balakrishnan2019visual,Jin_2018_CVPR,purohit2019bringing}, all of which are data driven under the deep learning framework. A large-scale motion dataset is required for the training of neural networks. Although the learning based methods maintain high inference efficiency, they suffer from the limitation of poor generalization on different types of motion. Besides, additional regularization is required to tackle the frame order ambiguity, which further increases complexity.

In this work, we report a generalized video extraction technique using the affine motion modeling. It enables to tackle multiple types of complex motion, offering an off-the-shelf tool to extract a high-speed video from a single blurred image. The main contributions are as follows.
\begin{itemize}
	\item We employ the affine transformation to model the rigid motion in three dimensions instead of the conventional two dimensions. This enables to not only reduce variable space, but also tackle various types of complex motion such as rotation and motion in depth.
	\item We introduce the differentiable affine operators including grid generator and grid sampler to realize gradient-descent optimization of the affine model. Besides, an additional $l0$-norm total variation regularization and coarse-to-fine enhancing strategy are also applied to attenuate artifacts and accelerate convergence.
	\item We apply stepwise affine transformation to the recovered reference image with the affine motion parameters, to produce the sharp video frames. The stepwise affine transformation maintains the nature to bypass the frame order ambiguity that commonly exists in dimension ascending tasks.
\end{itemize}

\begin{figure*}[t]
	\centering
	\includegraphics[width=1\linewidth]  {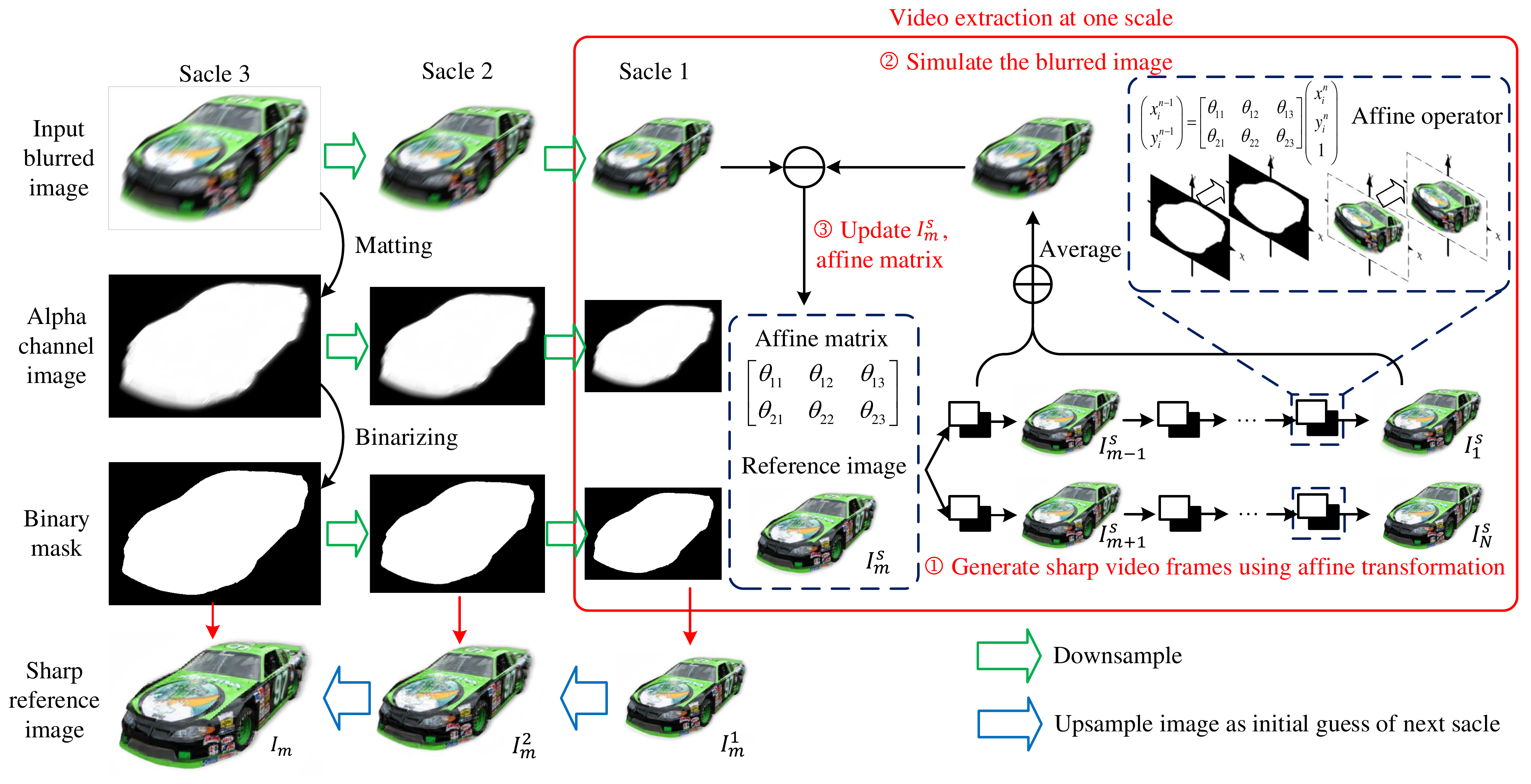}  
	\caption{Framework of the reported video extraction technique. With the input of a blurred image, different moving objects are first segmented for separate operation. For each object, with the initialized (or updated) reference image and affine parameters, the video frames are generated using the affine operators (grid generator and grid sampler). They are then averaged to produce the simulated blurred image. Both the reference image and affine parameters are updated via gradient-descent optimization following total variation and affine regularizations. This iterative optimization is applied for each moving object in different scales, to attenuate the ring artifact in a coarse-to-fine manner. Until convergence, the sharp video frames are retrieved by affine-transforming the reference image, without the frame order ambiguity.}
	\label{fig:framework}  
\end{figure*}

\section{Related Works}

\subsection{Blind Deblurring}

Blind motion deblurring has been extensively studied in the past few decades. The related techniques can be roughly categorized into Bayesian ones \cite{ruiz2015variational} and deep learning ones \cite{nah2019ntire}. Fundamentally, the Bayesian methods optimize the underlying sharp image by making it conform to the blur formation and statistical image priors, such as the mixture-of-Gaussian prior \cite{Fergus2006Removing}, the non-information uniform prior \cite{Levin2011Understanding}, the hyper-Laplacian prior \cite{amizic2012sparse}, the non-informative Jeffreys prior \cite{Babacan2012Bayesian}, the dark channel prior \cite{Pan2016Blind} and the 3D blur kernel prior \cite{whyte2012non}. Among these image priors, the total variation (TV) prior has been widely applied to improve reconstruction accuracy and attenuate the ringing effect. Kim et al. \cite{Kim2014Segmentation} introduced the TV-$l1$ model to estimate the sharp image and motion flow. Xu et al. \cite{Li2013Unnatural} developed an analytical solution for deblurring using the TV-$l0$ model. Pan et al. \cite{pan2016l_0} proposed the $l0$-norm gradient prior. Shao et al. \cite{shao2015bi} imposed the bi-$l0$-$l2$-norm regularization to further improve reconstruction quality. 

Since the Bayesian techniques suffer from high computational complexity, the data-driven deep-learning deblurring has gained more and more attentions. By engaging a large-scale dataset to train a neural network, the deep learning methods maintain fast inference speed \cite{DeblurGAN,Kupyn_2019_ICCV,nah2017deep,Sun2015CVPR}. Nah et al. \cite{nah2017deep} presented a multi-scale network to remove dynamic blur. Li et al. \cite{li2018learning} combined Bayesian model with a learning-based image prior to improve both reconstruction quality and efficiency. Different types of adversarial networks have been discussed for higher deblurring accuracy \cite{DeblurGAN,Kupyn_2019_ICCV}. Although the deep learning methods obtain high inference efficiency, it takes high cost for them to generalize on various types of motion as shown in the following experiment results.

Besides the above single-image deblurring methods, video deblurring has also been studied with additional temporal information. Kim et al. \cite{hyun2017online} introduced a temporal feature blending layer into an RNN to extract temporal features. Zhang et al. \cite{zhang2018adversarial} applied spatio-temporal 3D convolution in a CNN instead of RNN. Su et al. \cite{su2017deep} developed a U-shape decoder-encoder to remove camera shake blur. Ren et al. \cite{ren2017video} further exploited semantic features for auxiliary assessment of motion flow. In summary, video deblurring takes the advantage of spatio-temporal information, while single-image deblurring maintains temporal ambiguity.

\subsection{Video prediction}
Video prediction aims to predict more frames from a single or a few input images. Srivastava et al. \cite{srivastava2015unsupervised} proposed an unsupervised deep learning scheme for frame prediction. The LSTM-based encoder-decoder network are introduced to generate future frames. Mathieu et al. \cite{mathieu2015deep} applied GAN for video prediction, following the multi-scale architecture. Patraucean et al. \cite{patraucean2015spatio} and Finn et al. \cite{finn2016unsupervised} incorporated spatial transformer into networks to tackle per-pixel linear transformations instead of the globally consistent transformation. Vondrick et al. \cite{saito2017temporal,vondrick2016generating, 8237339} learned the semantic representation of videos using a generative adversarisal network, which enables to generate follow-up image sequences from one sharp image. Niklaus et al. \cite{jiang2018super,niklaus2017video} trained a convolutional network to estimate the motion flow between two given frames for intepolation. Bao et al. \cite{bao2019memc} took additional occlusion and contextual features into consideration to further improve reconstruction quality. 

We note that the video extraction task we are tacking is different from the video prediction problem. Although both the outputs are multiple sharp video frames, the input of video prediction is one or multiple sharp images. In video extraction, however, the input is a single blurred image. Fundamentally, video prediction predicts future frames, while video extraction tries to push the imaging speed limit and acquire ultra-fast dynamics.

\subsection{Video Extraction}

Video extraction from a blurred image was first demonstrated in 2018 by Jin et al. \cite{Jin_2018_CVPR}. They trained four convolutional networks, one of which is for the recovery of the middle frame, and each of the other three ones predicts two temporally symmetrical sharp frames. As a result, seven frames in total are retrieved from a single blurred image.
Purohit et al. \cite{purohit2019bringing} simplified the implementation of multiple networks by introducing the long short term memory (LSTM) units into a single recurrent video decoder.
Balakrishnan et al. \cite{balakrishnan2019visual} used a convolutional network to obtain prior motion features, which are then input into a deprojection network to recover sharp frames.
Although the above learning based methods maintain high inference efficiency, they suffer from the limitation of poor generalization on different types of motion. Besides, additional regularization is required to tackle the frame order ambiguity, which further increases complexity.

\section{Method}

In this work, we report a generalized affine-modeled video extraction technique that enables to tackle various 3D motion. The entire framework is presented in Fig. \ref{fig:framework}. With the single input blurred image, the reported technique contains the following four steps to retrieve sharp video frames.

\subsection{Segmentation in the alpha channel}

Considering that different objects follow different types of motion in one image, we first apply fore-background segmentation to separate different moving objects from background. Considering its high accuracy and strong generalization on various types of motion, we employ the closed-form matting technique \cite{levin2007closed} for segmentation and deriving the alpha channel image of each object. The alpha channel image $I_\alpha$ (ranging from 0 to 1) is a soft indicator map of foreground, which contains both location and motion information of the object.

\begin{figure}[h]
	\centering
	\includegraphics[width=1\linewidth]{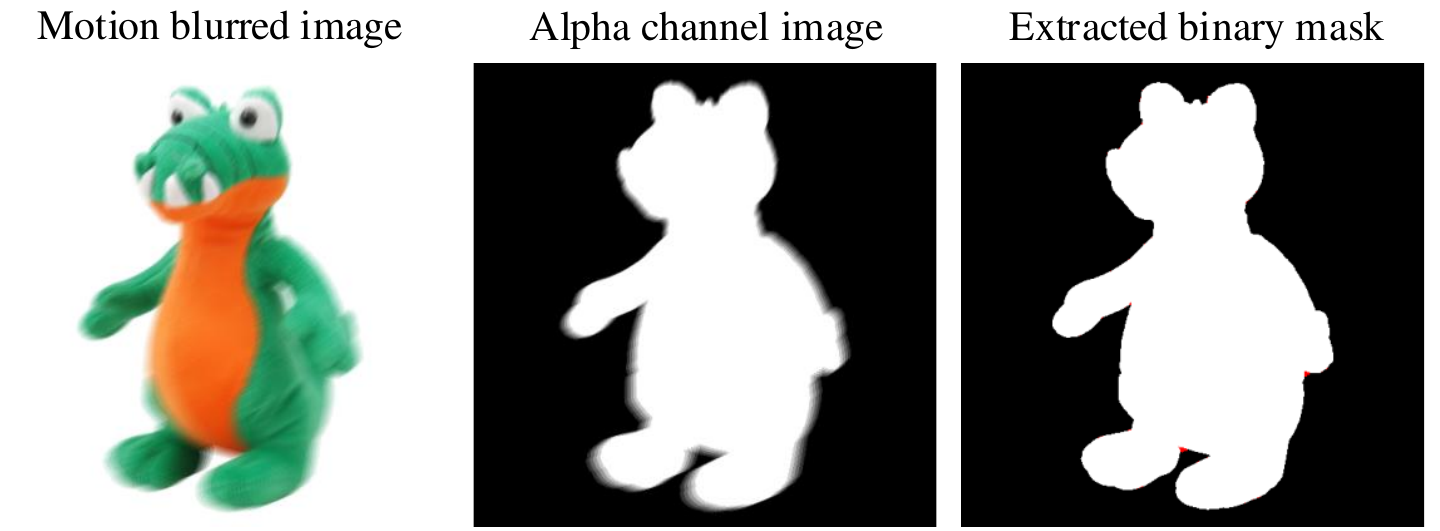}
	\caption{The alpha channel image and the binary mask of the middle frame from the motion blurred image. The motion blurred image (left) is synthesized with complex affine motion. The alpha channel image (middle) is derived by the close-form matting algorithm. We get the bianry mask (right) of the middle frame by directly rounding the alpha channel image. The red points in the right image indicate the difference between the gorund truth binary mask and the extracted one.}
	\label{fig:binarymask}
\end{figure}

With the alpha channel image, we further define the binary mask sequence of each object as $M_i~(i=1\to N)$ to indicate the object location in each video frame, where $N$ is the total number of video frames. The binary mask sequence follows $I_\alpha=\frac{1}{N}\sum_1^N{M_i}$, meaning that the alpha channel image is the temporal average of all the binary masks. Under the assumption of successive motion in one exposure, the binary mask of the middle video frame can be approximated by rounding the alpha channel image, while those of the other frames are derived by implementing affine transformation to the middle one. Fig. \ref{fig:binarymask} illustrates that the extracted binary mask has little difference from the ground truth binary mask. Both the alpha channel image and binary masks are employed to regularize the following solution of motion parameters.

\subsection{Affine modeling}

Conventionally, a motion blurred image $B$ is formulated by a common convolution model as $B = K\otimes I$,
where $K$ is a non-negative blur kernel representing the motion trajectory, and $I$ is the latent sharp image. Although this model has been widely applied for deblurring, it is limited to the specific motion that is parallel to the image plane, such as translation. When there exists an angle between the motion trajectory and imaging plane, the model fails to describe the blurring process. The study in ref. \cite{adiv1985determining} has demonstrated that when a rigid object moves, the position change at the image plane can be approximated by a 2D affine model, which is effective to reduce the parameter space. Accordingly, we reformulate the blurring formation using affine model, and denote
$A = \left[ \begin{matrix} \theta_{11}&\theta_{12}&\theta_{13}\\ \theta_{21}&\theta_{22}&\theta_{23} \end{matrix} \right]$ 
as the affine parameter matrix. The six parameters of $A$ control different types of motion, where 
$A_l = \left[ \begin{matrix} \theta_{11}&\theta_{12}\\ \theta_{21}&\theta_{22} \end{matrix} \right]$  
represents shape variation and rotation, and 
$A_t = \left[ \begin{matrix} \theta_{13} & \theta_{23} \end{matrix} \right]^T$  
controls the magnitude of translation. The affine matrix relates three successive sharp video frames ($I_{n-1},I_{n}$ and $I_{n+1}$) as
\begin{gather}
I_{n+1} = affine \left( I_{n}, A_l, A_t\right), 
\label{eq:affine1}
\end{gather}
\begin{gather}
I_{n-1} = affine \left( I_{n}, A_l^{-1}, -A_l^{-1}A_t\right),
\label{eq:affine2}
\end{gather}
Based on the denotations, the motion blurred image $B$ can be formulated as an average of a series of these sharp frames as 
\begin{gather}
B =f(I_m, A) = \frac{1}{N}\sum_{i=1}^{N}  \left(I_m + \sum_{i\neq m} affine \left( I_{m}, A \right)\right).
\label{eq:blur}
\end{gather}
Here we regard the middle frame $I_m$ as the reference image, and consequently reduce the variable space of $N$ images to one image with affine parameters. 

\begin{figure}[t]
	\centering
	\includegraphics[width=1\linewidth]{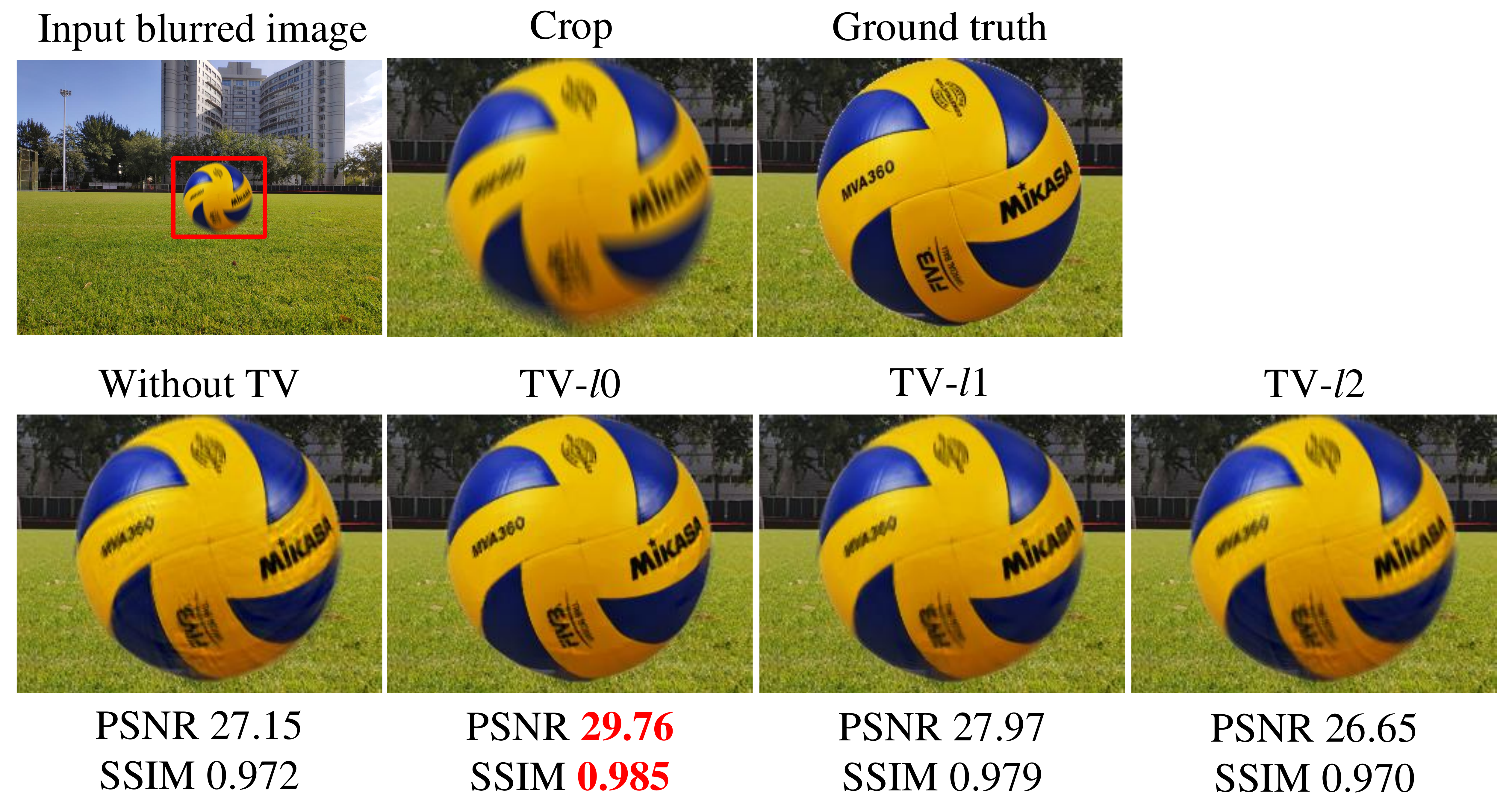}
	\caption{The reconstructed results using our method with different TV regularizations. The football follows hybrid motion containing translation and motion in depth. The second row is the extracted frame crops with different TV regularizations, which validate that the TV-$l0$ regularization maintains the highest reconstruction accuracy.}
	\label{fig:TV}
\end{figure}

To reconstruct the reference image and affine parameters, we formulate the objective function as
\begin{gather}
\min_{I_{m}, A} \left| f(I_m, A) - B \right| + \omega_{TV} TV\left(I_m\right) + p_{A}(A),
\label{eq:objective1}
\end{gather}
where the first data-residual term regularizes that the recovered video clip follows the blurred image formation. 

The second term $TV\left(I_m\right)$ is the total variation regularization \cite{rudin1992nonlinear} ($\omega_{TV}$ is a balancing weight), which arises from the prior that natural images are commonly smooth in brightness and have sparse variations. It helps attenuate the ring effect that degrades reconstruction quality. For each pixel $p$, it is defined as
\begin{gather}
TV\left(z\right)=\sum_{p} \phi\left(\partial_{*} z_p\right),
\end{gather}
where $\partial_{*} $ is the gradient operator and $*\in \left\lbrace h, v\right\rbrace $ stands for the horizontal and vertical directions.

We investigate different norms on total variation for the best performance, including the TV-$l0$ \cite{Li2013Unnatural}, TV-$l1$ and TV-$l2$. The $l0$-norm TV regularization refers to
\begin{gather}
\phi\left(\partial_{*} z_{i}\right)=\left\{\begin{array}{ll}
\frac{1}{\epsilon^{2}}\left|\partial_{*} z_{i}\right|^{2}, & \text { if }\left|\partial_{*} z_{i}\right| \leq \epsilon \\
1, & \text { otherwise }
\end{array}\right.
\label{TVl0}
\end{gather}
where $\epsilon$ is a small constant that gradually decreases from 1 to 0 in the iterative optimization process.
The reconstruction results of the middle frame using these different norms are shown in Fig. \ref{fig:TV}, from which we can see that the reconstruction without TV regularization contains serious artifacts. Among the multiple norms, the TV-$l0$ performs the best. Therefore, we employ TV-$l0$ in the following experiments.

\begin{figure}[t]
	\centering
	\includegraphics[width=1\linewidth]{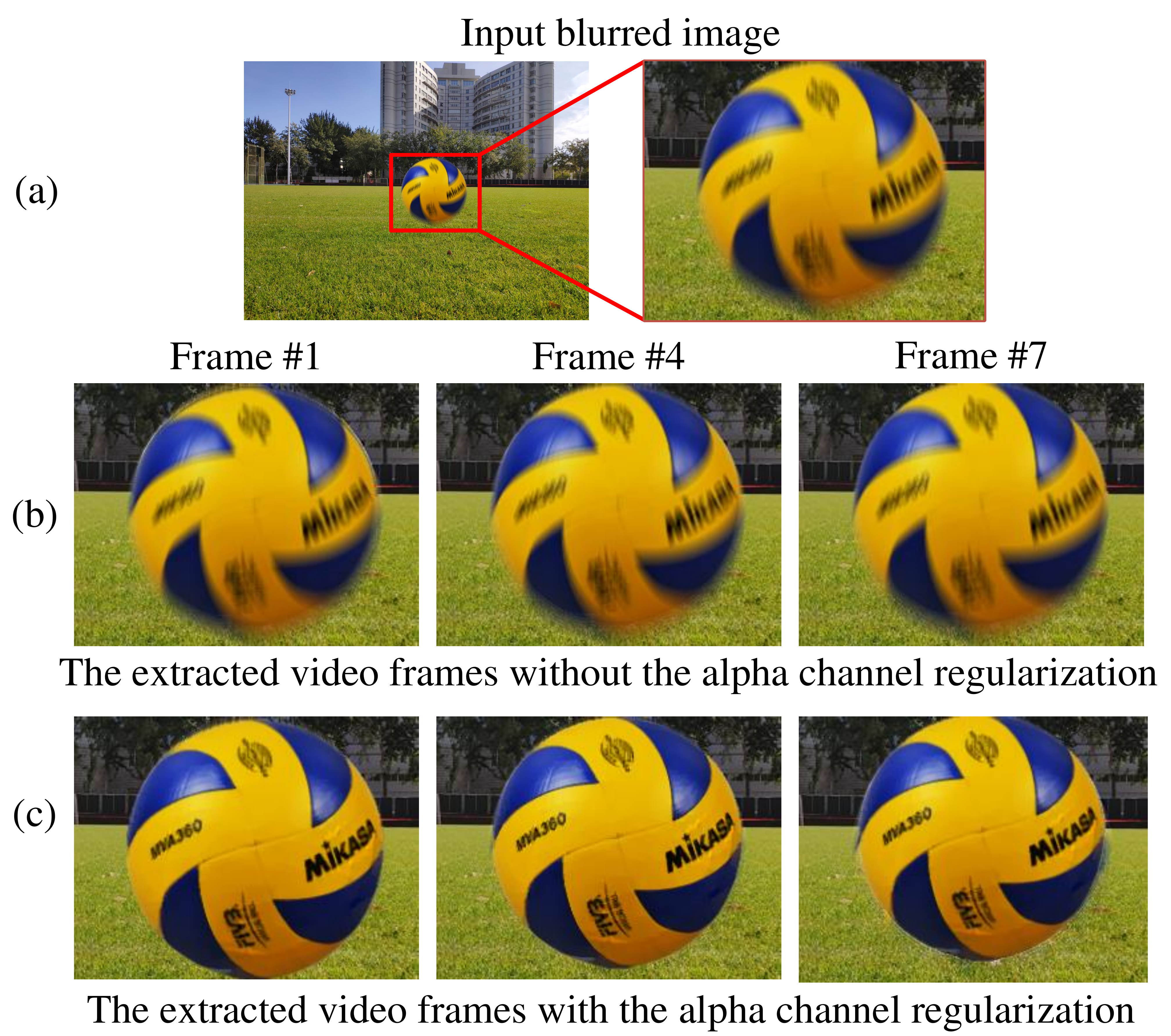}
	\caption{Demonstration of using the alpha channel regularization. (a) The input blurred image. (b) The extracted video frames without using the alpha channel regularization. (c) The extracted video frames using the alpha channel regularization.}
	\label{fig:alpha}
\end{figure}

The third term in Eq. (\ref{eq:objective1}) is the $l2$-norm affine matrix regularization
\begin{gather}
p_{A}(A) = \omega_l \| A_l - E \| _2^2 + \omega_t \| A_t\| _2^2 + \left| f(M_m, A) - I_{\alpha} \right|,
\label{eq:pA}
\end{gather}
where $E$ is a $2 \times 2$ identity matrix. This regularization is derived from the common sense that the motion between two successive frames is relatively little, and the affine matrix $[A_l, A_t]$ approximates to $[E, \bf{0}]$ \cite{marschner2015fundamentals}. The third term in Eq. (\ref{eq:pA}) is the alpha channel regularization which further helps increase recovery accuracy of the motion parameters in $A$. We evaluate the reconstruction results with and without employing this constraint, and the results are shown in Fig. \ref{fig:alpha}. We can see that the reconstructed frames without the alpha channel regularization contains blur, while those using this regularization are more clear and sharp. We can clearly observe the football's motion and shape varization, which validates the effectiveness of the alpha channel regularization.

\subsection{Affine Optimization}

As the variable space has been reduced to $I_m$ and $A$, we derive an affine optimization algorithm to solve the above model in Eq. (\ref{eq:objective1}). Following the variable splitting strategy \cite{Li2013Unnatural}, the objective function is decomposed into two alternating sub-functions regarding $I_m$ and $A$ respectively as
\begin{gather}
\min_{I_m}  \left| f(I_m, A) - B \right| + \omega_{TV} TV\left(I_m\right), ~\text{with fixed} \ A,
\label{eq:subobj1}
\end{gather}
\begin{gather}
\min_{A}  \left| f(I_m, A) - B \right| + p_{A}(A), \text{with fixed} \ I_m.
\label{eq:subobj2}
\end{gather}

The conventional optimization methods are not applicable for the above affine model, because the employed affine transformation is based on coordinate calculation instead of pixel calculation \cite{jaderberg2015spatial}. Inspired by the spatial transformer network \cite{jaderberg2015spatial}, we introduce the differential affine operators to enable the gradient-descent optimization of the affine model.
The differential affine operators include a grid generator and a grid sampler. The grid generator produces the affine-transformed coordinate, and the grid sampler outputs the transformed image under this coordinate. Mathematically, assuming that the coordinate of the input image is $(x^s, y^s)$ and the transformed one is $(x^t, y^t)$, the point-wise affine transformation is defined as
\begin{gather}
\left(\begin{array}{l}
{x_{i}^{s}} \\
{y_{i}^{s}}
\end{array}\right)=A\left(\begin{array}{l}
{x_{i}^{t}} \\
{y_{i}^{t}} \\
{1}
\end{array}\right)
=\left[\begin{array}{lll}
{\theta_{11}} & {\theta_{12}} & {\theta_{13}} \\
{\theta_{21}} & {\theta_{22}} & {\theta_{23}}
\end{array}\right]
\left(\begin{array}{l}
{x_{i}^{t}} \\
{y_{i}^{t}} \\
{1}
\end{array}\right),
\label{eq:coordinate}
\end{gather}
where $A$ is the affine parameter matrix, and the subscript $i$ denotes the pixel index.

With the affine-transformed coordinate, a grid sampler is used to calculate the affine-transformed image \cite{jaderberg2015spatial}. Specifically, denoting $I^{nm}_{s}$ as the pixel value at location $(n,m)$ of the input image, the pixel value $I^i_{t}$ of the transformed image at the $i$-th pixel location is a weighted summation of all the pixels as \cite{jaderberg2015spatial}
\begin{gather}
I^i_{t}=\sum_{n}^{H} \sum_{m}^{W} I^{n m}_s \max \left(0,1-\left|x_{s}^{i}-m\right|\right) \max \left(0,1-\left|y_{s}^{i}-n\right|\right),
\label{eq:model}
\end{gather}
where $H$ and $W$ stand for the image's height and width. Equation (\ref{eq:model}) shows that the differential grid sampler copys the value at the nearest pixel $\left(x_{s}^{i}, y_{s}^{i}\right)$ to the output location $\left(x_{t}^{i}, y_{t}^{i}\right)$. Based on the above denotations, the partial derivatives of the output pixel $I^i_t$ with respect to $I^{n m}_s$ and $x_{s}^{i}$ are derived as \cite{jaderberg2015spatial}
\begin{gather}
\frac{\partial I^i_{t}}{\partial I^{nm}_s}=\sum_{n}^{H} \sum_{m}^{W} \max \left(0,1-\left|x_{s}^{i}-m\right|\right) \max \left(0,1-\left|y_{s}^{i}-n\right|\right),
\label{eq:gradient1}
\end{gather}
\begin{gather}
\frac{\partial I^i_{t}}{\partial x_{s}^{i}}=\sum_{n}^{H} \sum_{m}^{W} I^{n m}_{s} \max \left(0,1-\left|y_{s}^{i}-n\right|\right)g(x_s^i, m),
\label{eq:gradient2}
\end{gather}
where
\begin{gather}
g(x_s^i, m)=\left\{\begin{array}{ll}
0, & \text { if }\left|m-x_{s}^{i}\right| \geq 1 \\
1, & \text { if } m \geq x_{s}^{i} \\
-1, & \text { if } m<x_{s}^{i}.
\end{array}\right.
\end{gather}
$\frac{\partial I^i_{t}}{\partial y_{s}^{i}}$ follows the similar derivation as Eq.  (\ref{eq:gradient2}).

\begin{figure}[t]
	\centering
	\includegraphics[width=1\linewidth]{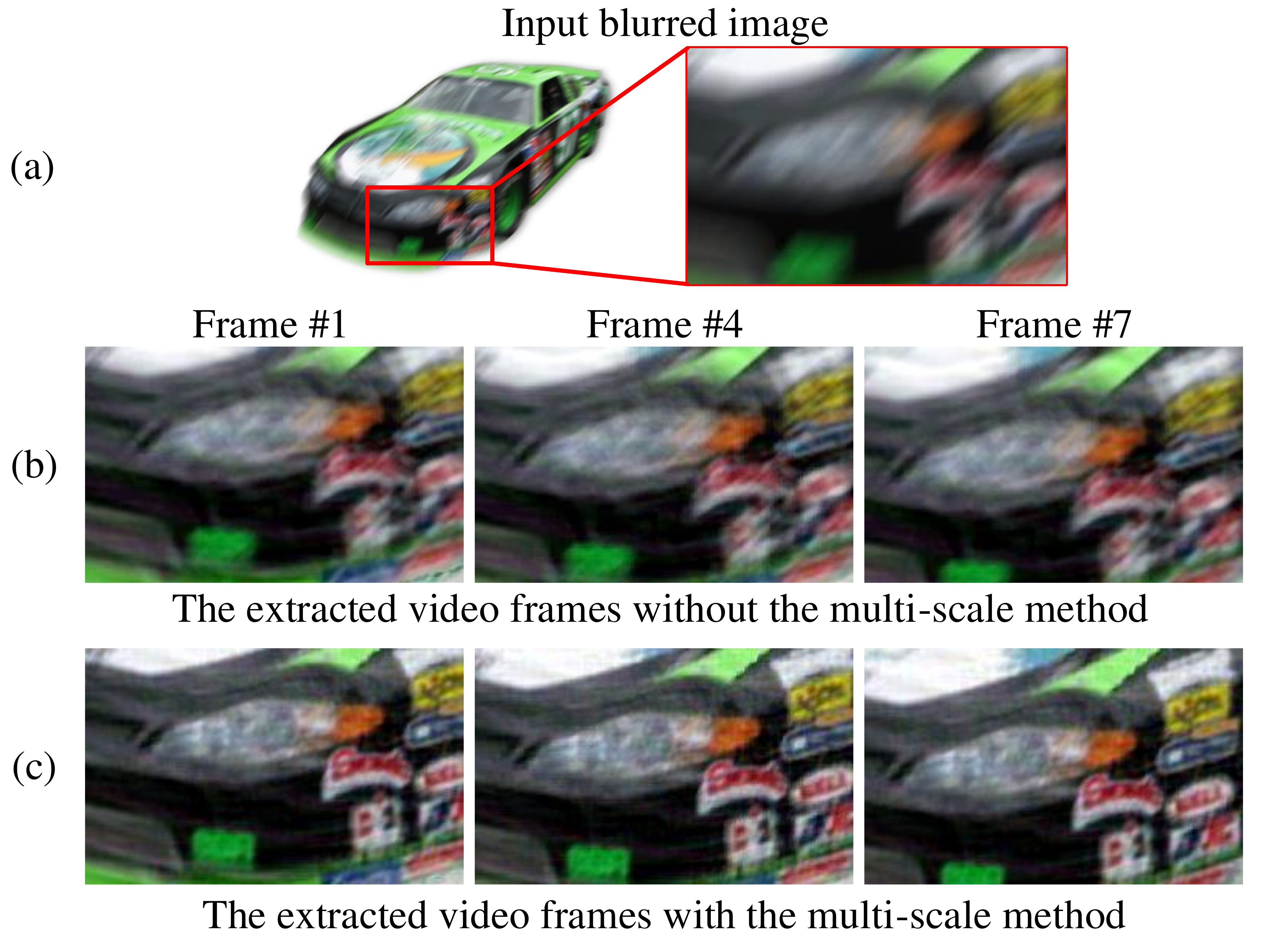}
	\caption{Demonstration of using the coarse-to-fine enhancing strategy for removing artifacts. (a) The input blurred image from Dai et al.'s dataset \cite{dai2008motion}. (b) The extracted video frames without using the coarse-to-fine enhancing strategy. (c) The extracted video frames using the coarse-to-fine enhancing strategy.}
	\label{fig:multi}
\end{figure}

Based on the above differentiable sampling mechanism, by grouping the pixel-wise derivative of Eq. (\ref{eq:gradient1}) into Eq. (\ref{eq:subobj1}), we realize the gradient-descent updating of the reference image $I_m$. Similarly, we relate the derivatives of Eq. (\ref{eq:gradient2}) with Eq. (\ref{eq:subobj2}), and enable the loss gradient to flow back to the sampling grid coordinates, and therefore back to the affine parameter matrix $A$ through Eq. (\ref{eq:coordinate}). In such a strategy, the objective model in Eq. (\ref{eq:objective1}) becomes solvable in a gradient descent manner.

\subsection{Coarse-to-fine enhancing} 
In order to further attenuate artifacts and improve reconstruction accuracy, we develop a coarse-to-fine optimization scheme for our algorithm. Recent research validates that the coarse-to-fine scheme enables to reduce ring artifacts \cite{xiao2015stochastic}. As shown in Fig. \ref{fig:framework}, the optimization process is divided into three scales. At the coarsest scale, in order to reconstruct a low-resolution intermediate result, the blurry image, alpha channel image and binary mask are first downsampled to half of the original resolution. When the latent reference image of low resolution is recovered, it is then bicubic upsampled at a factor of $\sqrt{2}$ as a initial guess for the next scale. Considering that the affine matrix $A$ is normalized by image size, it can be directly employed for the next scale.

We test video extraction accuracy with and without using the coarse-to-fine strategy. From the results shown in Fig. \ref{fig:multi}, we can see that the extracted video frames without using the coarse-to-fine strategy still contain motion blur, while those using the coarse-to-fine strategy are more sharp and clear. The results validate that the coarse-to-fine scheme enables to retrieve sharp video frames. Further, by comparing the video frames at different moments, we can clearly see the retrieved motion.

\begin{algorithm}[]
	\SetKwInOut{Majorization}{Majorization}\SetKwInOut{Minimization}{Minimization}
	\SetKwData{set}{set}
	\SetKwInOut{Initialization}{Initialization}\SetKwInOut{Input}{Input}\SetKwInOut{Output}{Output}
	\vspace{2mm}
	\Input{Motion blurred image $B$, alpha channel image $I_{\alpha}$, frame number $N$, numbers of iteration $T$.}
	\Output{Reconstructed frames ${I_1,I_2,\dots,I_{N}}$.}
	\vspace{2mm}
	Initialization: $I_{fm}=\bf{0}$, $I_{bm}=\bf{0}$, $A_l$ is set to be a random $2\times 2$ matrix, $A_t=[0,0]^T$\;
	
	Solve the binary mask of middle frame: $M_m=round(I_{\alpha})$ \;
	\For{$s = 1 \to 3$}{
		$sacle = (\sqrt{2})^{s-3}$ \;
		$B^s=downsample(B, scale, 'bicubic')$ \;
		$I_{\alpha}^s=downsample(I_{\alpha}, scale, 'bicubic')$ \;
		$M_m^s=downsample(M_m, scale, 'bicubic')$ \;
		\If{$s>1$}{ 
			$I_f^s=upsample(I_f^{s-1}, \sqrt{2}, 'bicubic')$\;
			$I_b^s=upsample(I_b^{s-1}, \sqrt{2}, 'bicubic')$\;
		}
		\For{$t=1:T^s$}{
			update $I_m^s$ with fixed $A$ following Eq. (\ref{eq:subobj1})\;
			update $A$ with fixed $I_m^s$ following Eq. (\ref{eq:subobj2})\;
		}
	} 
	
	Retrieve sharp frames $I_1, I_2, ..., I_N$ following Eqs. (\ref{eq:affine1},\ref{eq:affine2}) with the optimized $I_m$ $A_l$ and $A_t$.
	\caption{\small{The affine-modeled video extraction algorithm.}}
	\label{alg}
\end{algorithm}

\begin{figure*}[t]
	\centering
	\includegraphics[width=0.9\linewidth]{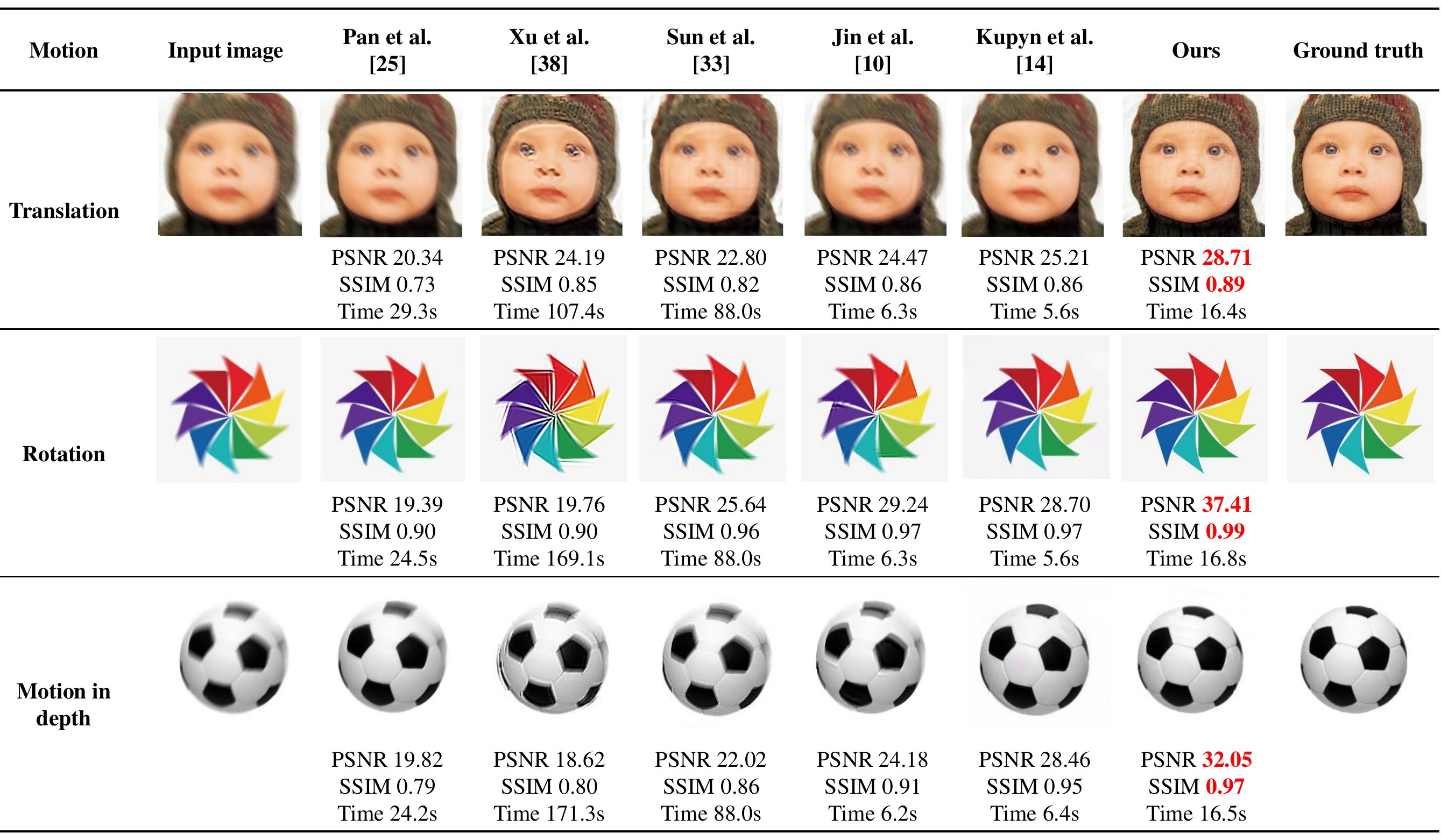}
	\caption{The comparison of different methods for middle frame reconstruction under different types of motion, including translation, rotation and motion in depth. Both the qualitative and quantitative results are presented, which validate the strong generalization ability and state-of-the-art performance of the reported technique.}
	\label{fig:motion}
\end{figure*}

We summarize the workflow of the reported video extraction technique as follows (Alg.\ref{alg}). We first segment the moving foreground object in the alpha channel and derive binary masks from the alpha image. As for each moving object,  we formulate the affine motion blur model, and employ TV-$l0$  and alpha channel priors as regularization to form the objective function. To enable gradient-based optimization of the affine model, we introduce the differential affine operator, and apply the coarse-to-fine enhancing strategy to further remove artifacts and improve reconstruction accuracy. As the iteration converges, we obtain the recovered sharp reference image $I_m$ and affine parameter matrix $A$. The sharp video frames at the other moments can be conveniently reconstructed following Eq. (\ref{eq:affine1}) and (\ref{eq:affine2}). 

\begin{figure*}[t]
	\centering
	\includegraphics[width=1\linewidth]{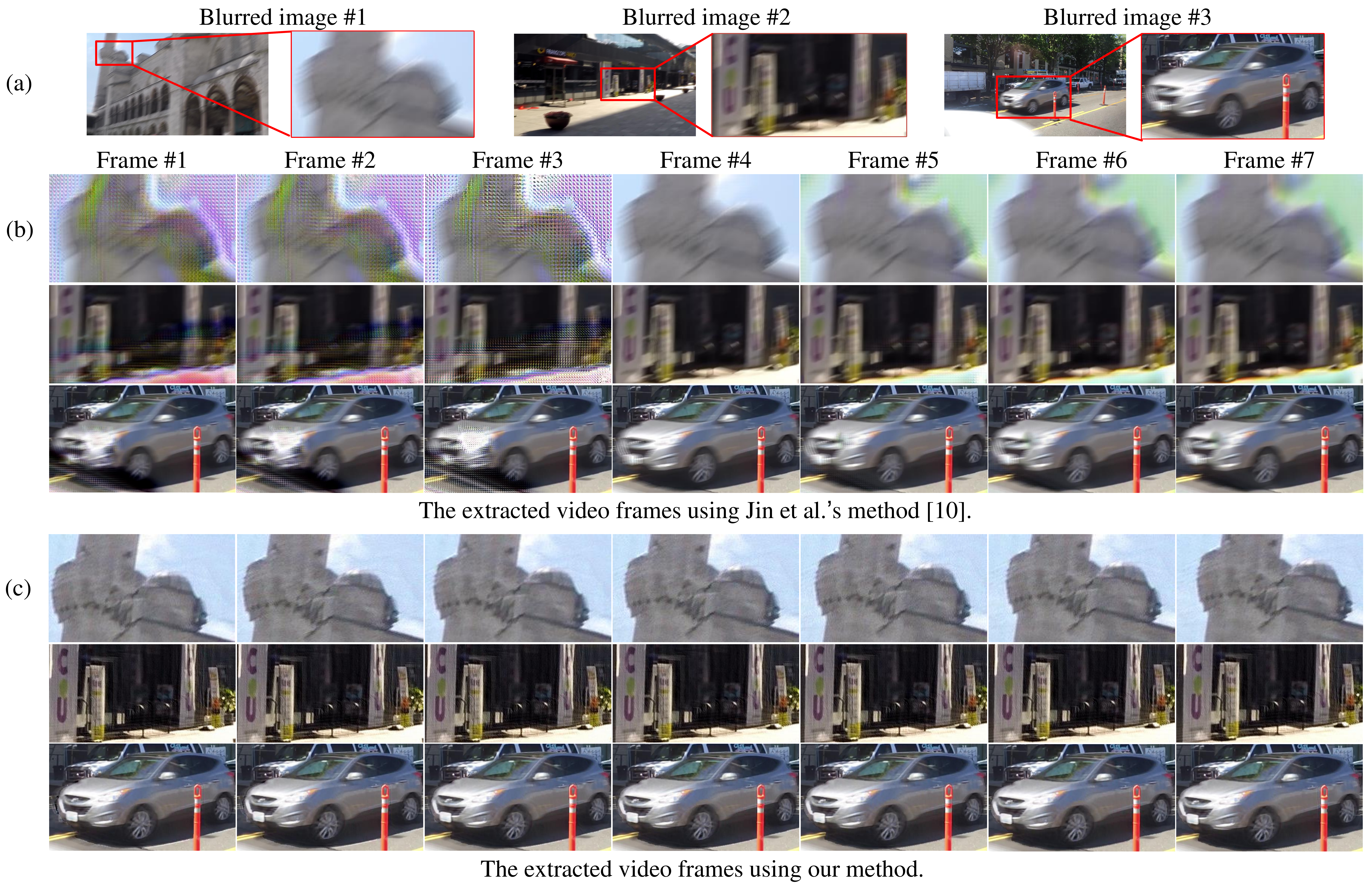}
	\caption{The exemplar retrieved video frames from Nah et al.'s \cite{nah2017deep} and Su et al.'s public datasets \cite{su2017deep}. (a) The input blurred images. (b) The extracted video frames by Jin et al.'s network \cite{Jin_2018_CVPR}. (c) The extracted video frames by our method.}
	\label{fig:dataset}
\end{figure*}

\section{Experiments}

In the next, we perform a set of experiments to validate the superiority of the reported technique over the existing methods. We set the regularization weights in Eq. (\ref{eq:objective1}) as $\omega_{TV}=1\times 10^{-9}, \omega_{\alpha}=0.3, \omega_l=10$ and $\omega_t=1$, which perform best among a number of simulation trials. The learning rates for optimizing the reference image and affine parameters are set 0.02 and 0.01, respectively. The iteration numbers are set $50,100,150$ for the three scales in the coarse-to-fine optimization scheme. The relaxation parameter $\epsilon$ in Eq. (\ref{TVl0}) is initialized as 1, and it is divided by 2 for every 50 iterations.
The number of recovered video frames $N$ is preset by users. We note that it is correlated to the motion range of the input blurred image. Empirically, a bigger $N$ is appropriate for wider motion range, which ensures the validation of the little motion prior between two successive frames. In the following experiments, we set $N=7$ for a fair comparison with the state-of-art method \cite{Jin_2018_CVPR} that can only produce 7 frames.

For the evaluation of reconstruction quality, we employ two quantitative metrics including the peak signal to noise ratio (PSNR) and the structural similarity metric (SSIM) \cite{wang2004image}. PSNR measures the overall intensity difference between two images, and SSIM evaluates the similarity of structural details.
Our algorithm is implemented using Python with an NVIDIA GTX-1060 GPU.

\subsection{Middle frame deblurring on different motion types}

To validate the generalization of the reported technique on different types of motion, we apply it to solve three common types of motion including translation, rotation and motion in depth. The input blurred images are synthesized using different blur kernels on sharp images. To compare with the deblurring methods, the quantitative metrics are calculated using the recovered middle reference frame for our method and Jin et al.'s method \cite{Jin_2018_CVPR}. The deblurring methods \cite{Pan2016Blind} and \cite{Li2013Unnatural} is based on the TV-$l0$ prior, and \cite{Pan2016Blind} is a uniform-deblurring algrithom while \cite{Li2013Unnatural} focuses on non-uniform blur. \cite{Kupyn_2019_ICCV} and \cite{Sun2015CVPR} are learning-based methods. Kupyn et al's method \cite{Kupyn_2019_ICCV} maintains the state-of-the-art deblurring performance. Both the qualitative and quantitative results are presented in Fig. \ref{fig:motion}, where the running time of each method is also provided (tested under the image size of 256$\times$256). Among the existing deblurring methods, Kupyn et al's network \cite{Kupyn_2019_ICCV} produces the best reconstruction performance. However, for the motion in depth, it produces a sharp image with clear edges rather than the real shape of the moving object. As a comparison, the reported technique produces the highest reconstruction accuracy for different types of motion. Although Jin et al.'s method employs deep learning for fast inference speed, it fails for most of the cases with weak generalization.

\begin{figure*}[t]
	\centering
	\includegraphics[width=0.9\linewidth]{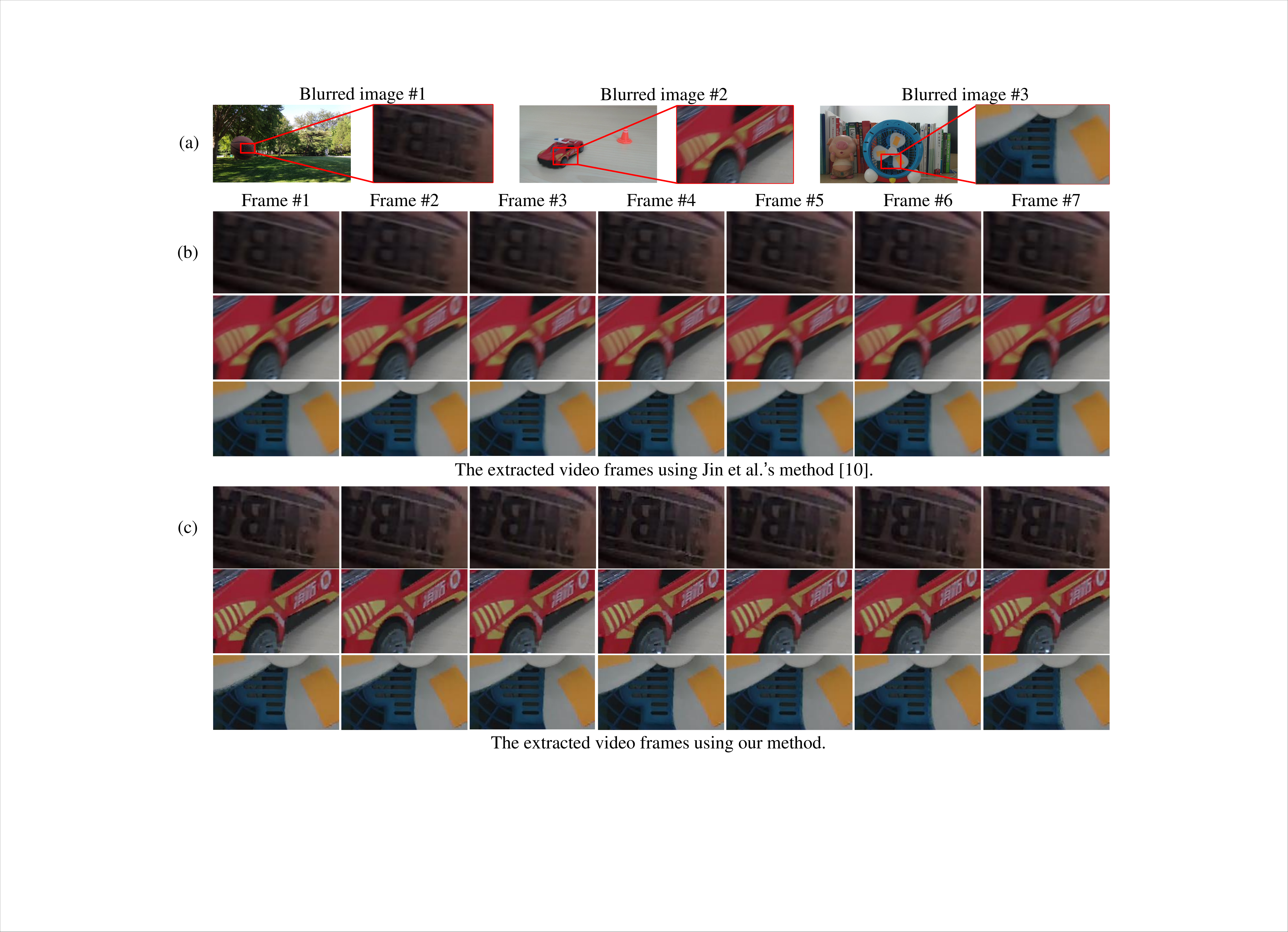}
	\caption{The retrieved video frames of real captured blurred images (translation motion / motion in depth / rotation). The blurred images are captured by an Andriod smartphone at 30 fps. (a) The captured blurred images as input. (b) The extracted video frames by Jin et al.'s network \cite{Jin_2018_CVPR}. (c) The extracted video frames by our method.}
	\label{fig:real}
\end{figure*}

\subsection{Testing on public datasets}

To further validate the generalization of our method on different scenes, we test it on public datasets including Nah et al.'s dataset \cite{nah2017deep} and Su et al.'s dataset \cite{su2017deep}. The sharp frames of two datasets are captured at 240 frames per second. The blurred images $\#$1 and $\#$2 from Nah et al.'s dataset \cite{nah2017deep} are simulated by averaging sucessive 15 frames while the blurred image $\#$3 from Su et al.'s dataset \cite{su2017deep} is synthesized by averaging 5 sharp frames. 
The references \cite{balakrishnan2019visual, Jin_2018_CVPR, purohit2019bringing} are the most recent papers for video extraction. However, the code from Jin et al \cite{Jin_2018_CVPR} is the only public available demo code.
The exemplar retrieved video frames are shown in Fig. \ref{fig:dataset}, where the image $\#$1 and $\#$2 are with global motion blur, and the image $\#$3 is with local blur of motion in depth. For image $\#$1, we can see that the retrieved frames of Jin et al's method suffer from severe color distortions. For image $\#$2 and $\#$3,  Jin et al's method still produces blur and artifacts. In comparison, the reported technique enables to recover sharp characters from the blurred image $\#$2 and clear textures from the blurred images $\#$1 and $\#$3. Further, by comparing the video frames at different moments, we can clearly see the retrieved motion. The retrieved shape variation of the moving car in image $\#$3 validates that our algorithm can effectively recover 3D motion.  To conclude, the above results validate the superiority of the reported technique with enhanced reconstruction quality and visual performance.

\subsection{Results on real captured data}

To demonstrate the reported technique in practical applications, we used an Andriod smartphone (Xiaomi Mi 8) to acquire motion blurred images, and applied our method and Jin's algorithm to retrieve video frames. The input images are captured at 30 frames per second, with the size of 1920$\times$1080 pixels. The image $\#$1 corresponds to translation motion. We threw the basketball parallel to the imaging plane. Compared to the distance between basketball and camera, the motion in depth of the basketball is slight that can be ignored. The image $\#$2 corresponds to the motion in depth. We pushed the toy car moving from near to far at a certain angle. The image $\#$3 corresponds to rotation. We electrified the fan and captured the motion of rolling blades.
Fig. \ref{fig:real} presents the retrieved video frames from these real captured blurred images. For the blurred image $\#$1 and $\#$2, our method can retrieve sharp characters on the moving foreground, while the extracted frames of Jin's method still contain blur. For the blurred image $\#$3, Jin's method fails to recover the rotation motion of the fan's blade. In comparison, our method reconstructs sharp video frames with clear motion. To conclude, the reported technique enables to recover sharp video frames from real captured motion blurred images, which validates its effectiveness in practical applications.

\section{Conclusion and discussion}
In this work, we report a generalized affine optimization technique, enabling to retrieve a high-speed video from a single blurred image. Its innovations lie in the following three aspects. First, we model the to-be-retrieved multiple video frames by a sharp reference image and its affine transformations. This enables to reduce variable space, and tackle various types of complex motion such as rotation and motion in depth. Second, we decomposite the blurred image into  parts of different motion status by segmentation. Each part of sharp reference image is estimated seperately. What's more, the alpha channel segmentation is exploited as a prior to enhance estimation accuracy of motion parameters. Third, we realize gradient-descent optimization of the affine motion model by introducing the differential grid operators. Further, we introduce the coarse-to-fine scheme to attenuate ring effect. The multiple sharp video frames are finally reconstructed by the stepwise affine transformation of the reference image, which maintains the nature to bypass the common frame order ambiguity. Experiments on both public datasets and real captured data validate the state-of-the-art video extracting performance of the reported technique.

The technique provides a novel and generalized framework to extract high-speed video frames from a single blurred image. It can be further investigated to improve performance in the following several aspects. First, to tackle the spatially non-uniform motion in the case of multiple objects,  each object can be processed individually by parallel computing for high efficiency.  Second, to remove the negative effect of poor segmentation on reconstruction quality, we can further use the per-pixel affine transformation to estimate the motion of each object. Third, although the assumption of temporally uniform motion over the exposure time is valid for a majority of high-speed applications, there still exist a few cases when the motion is not temporally uniform. To tackle this problem, we can further divide the exposure time into multiple periods, and employ different affine parameters to model the motion in each period. In summary, the reported technique can be widely extended. It is interesting and worth to further investigate the technique's extension for various practical applications.


%


\appendices




\ifCLASSOPTIONcaptionsoff
  \newpage
\fi



\bibliographystyle{IEEEtran}
\bibliography{egbib.bib}
\end{document}